# A Self-Adaptive IoT-based Approach for Improving the Decision Making of Active Surgical Robots in Hospitals

Alina Saduova[1], and Eyhab Al-Masri[1*]
[1]School of Engineering and Technology, University of Washington Tacoma
*Corresponding author: ealmasri@uw.edu

**Abstract**

In recent years, surgical robots have become instrumental tools for assisting surgeons in performing complex surgical procedures in hospitals. Unlike conventional surgical methods, robotic systems help surgeons, for example, to perform minimally invasive surgical procedures while enhancing the precision and control of operations (e.g. tiny incisions, wound sutures, endoscopic suturing, among others). To this extent, it is essential to consider several factors that may influence the feasibility and decision making of employing robotic systems in surgical procedures. In this paper, we propose an IoT-based self-adaptive approach that uses multi-criteria decision analysis methods (MCDA) for enhancing the decision making of operations involving surgical robots. Throughout this paper, we present experimental validation results in utilizing MCDA as an effective strategy for enhancing the decisions of employing robotic systems in surgical procedures.

## Introduction

In recent years, the Internet of Things (IoT) has brought about new opportunities across many disciplines for allowing objects to interact with each other [1]. The area of medical surgeries is no exception. IoT technology enables heterogeneous robots to interact together during medical surgeries for improving surgical outcomes. Traditionally, at the surgical level, robots remained passive such that they serve as assistive tools during some surgical operations. The problem is, however, that the outcome of such operations depends primarily on the skill level of the surgeon and the surgical training or assessment methods in the operating room, which can often be time consuming and subjective [2]. To this extent, identifying an effective mechanism for improving the self-adaptive feedback control for surgical robots becomes essential.

Through an IoT-based approach, it is then possible to combine deeper and richer information of the robot's contextual situation during surgical procedures for helping surgeons during their assessment when performing surgical procedure involving robots. Robots have been widely adopted across medical areas for performing a wide range of tasks related to orthopedic, urologic and neuro surgeries. Unlike passive robots, advancements in the area of robotic systems allowed a more active role of robots as being assistive tools that can help in the decision-making process or even make their own decisions, independently.

Among the first semi-active robot-assistive surgeries are the ones that took place in Cleveland (USA) using ZEUS, a three-armed robot that was controlled by surgeons to perform laparoscopic fallopian tube anastomosis [3]. Around the same time ZEUS was introduced, the world's first ever robot-assistive heart bypass surgery that was performed in Germany using the da Vinci Robot, one of the most commonly known robot for assistive surgeries [4]. Unlike open surgeries, using robots during minimally invasive surgeries reduces the risk of bodily injuries. Generally, there are three main types of surgical robots employed in minimally invasive surgeries including: (a) passive robots, (b) semi-active and (c) active robots.

Passive robots, which are commonly hand-held devices, are assistive tools that do not actively perform any surgical operations. An example of passive robots is NeuroMate, a commercial robot that helps in locating large brain tumors in the brain and integrates to a planner in neurosurgery [5]. Semi-active robots are ones that allow surgeons to perform a surgical procedure having direct control. That is, semi-active robots can interact directly on the patient's body with the direct control of an observer or surgeon (e.g. Acrobot Surgical System [6]). Active robots are autonomous and are capable of performing part of surgical procedures (e.g. ROBODOC [7], CASPAR [8]).

The main goal of active robots is not to replace surgeons but rather provide them with active instrumentation equipped with powerful features for assisting them during surgical procedures. Generally, active robots support teleoperation such that surgeons remotely control them outside of an operating room. Active robots provide surgeons with a greater level of flexibility and control. In this paper, we primarily focus on active surgical robots. The objective of this research effort is to consider contextual information collected by the robot to enhance the decision-making by surgeons. We follow an IoT-based approach such that deeper and richer contextual information collected are considered.

We believe that by employing an IoT-based approach, it is then possible for robots to play an increasingly active role in robotic surgeries while considering a risk factor when they make decision during surgical operations or procedures. To this extent, we employ a multi-criteria decision analysis (MCDA) method for enhancing the decision making of robotic assistive surgeries. By considering factors and constraints related to real-time captured events, robots can help surgeons in making accurate steps during surgeries. We primarily focus on three key factors: (a) collecting contextual information, (b) determining the risks associated with every computed robot trajectory plan during surgery and (c) recommending a plan for surgeons.

The rest of this paper is organized as follows. Section 2 introduces the IoT-based Surgical Robot (IoTSR) framework, a middleware used for improving the decision making of active surgical robots. Section 3 presents the results and evaluation of our proposed optimization strategy of the IoTSR framework. Finally, Section 4 provides conclusion and future work.

## An Edge-based Surgical Robot (IoTSR) Framework

To improve the decision making of active surgical robots, we present an edge-based self-adaptive Surgical Robot (IoTSR) framework acting as a middleware between patients, surgical robot in an operating room and a surgical team or experts as presented in Figure 1.

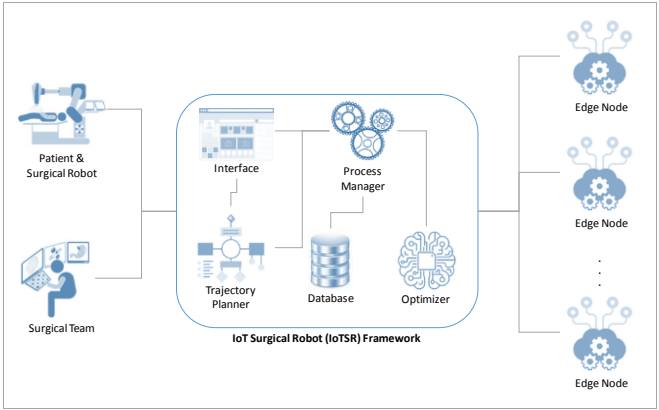

Fig. 1 Edge-based IoT Surgical Robot (IoTSR) Framework.

IoTSR is designed as an edge-based system such that it can reside within a hospital environment and process incoming tasks across multiple edge computing nodes. This allows IoTSR to offload edge-based tasks that are computationally intensive (e.g. AI operations) across other distributed edge nodes discovered within a hospital's network. Using the built-in sensors, the surgical robot continuously collects data about the patient. This data primarily depends on the operation that is performed. For example, an active surgical robot can be programmed to perform very specific procedures such as drainage of hepatic duct, right-hepatic vessel dissection, hepatic vascular control, among others [9]. Therefore, it is imperative to design IoTSR such that it supports multiple-criteria for the decision making process.

The process manager processes incoming requests from and to the surgical robot and surgical team. The surgical robot continuously transmits data via a local network to the IoTSR which is then stored into a database. Depending on the request, IoTSR processes the requests through an optimizer component that uses an optimization strategy for identifying a possible solution. This optimization strategy depends primarily on the attributes and factors considered as part of robotic surgery procedure.

The optimizer is designed such that it is self-adaptive and is continuously improving its process over time. The trajectory planner uses a multi-criteria decision method (MCDM) to recommend an action to the surgical team. This enables the IoTSR to provide assistance to the surgical team and improves the confidence level for actions to be performed in a real-time manner. As part of the optimization strategy, IoTSR considers a number of decision variables that are procedure-dependent. For example, for a hepatic vessel dissection procedure, IoTSR allows surgeons to identify decision variables that play a vital role in the decision making such as bleeding rate or blood loss, camera visual clarity of a camera, risk rate of vessels to cancer, relevance to liver, among others.

The decision variables are part of the trajectory planner component. We assume that the values for the decision variables are measured or can be computed by the robot (e.g. sensors measurements or probability models). Decision variables can be associated with one or more constraints that can be defined by a surgical team. For example, when considering the estimated blood loss (EBL) rate, a surgeon can use IoTSR to specify the threshold for blood loss rate or minimum allowable percentage of a visual clarity factor via a camera.

The constraints play a major role in the overall decision-making process and provides surgical teams flexibility to predefine their own preferences as to how the surgical robot can be utilized. IoTSR provides an interface such that surgeons can specify their own preferences and decision variables' constraints. The trajectory plan is formulated based on the decision variables including an objective function. A decision matrix is then computed. We employ the Technique for Order Preference by Similarity to Ideal Solution (TOPSIS) decision method for determining the best or most suitable choice to be performed during a surgical operation. Figure 2 presents an overview of IoTSR's decision-making process using TOPSIS. Algorithm 1 presents a pseudo-code algorithm for IoTSR's decision-making process and trajectory plan optimization.

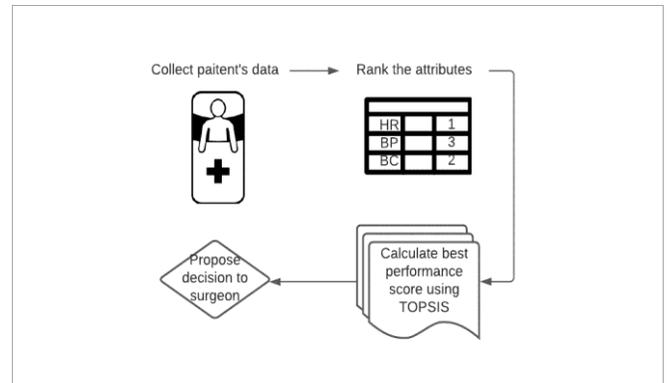

Fig. 2 IoTSR Decision-Making Process using TOPSIS.

---

**Algorithm 1** IoTSR Trajectory Plan TOPSIS Optimization

1: **procedure** optimizeTrajectoryPlan
2:   $X \leftarrow Null$
3: calculate normalized matrix ($X¬ij$) for all decision variables
4: **while** true **do** // for all trajectory plans
5:   **for** each decision variable
6:     calculate normalized weighted value
7:   calculate performance value (**RScore**) for each trajectory plan
8:   **if** $P_{max} <$ RScore:
9:     set $P_{max}=$ RScore
10: **return** $P_{max}$ value and the corresponding trajectory

---

## Results and Evaluation

One of the main goals of the IoTSR framework is to assist surgeons by improving decision-making process when performing precise robot-assisted surgical procedures while reducing the procedural time. Because there may exist a wide range of possible surgical operations that can be conducted using robots, we limited the scope of our evaluation to a surgical plan that involves pancreaticoduodenectomy (also called the Whipple procedure) [11, 12] with an extended hepatectomy (or liver resection) [10]. Each procedure can have different decision variables, which requires IoTRS to flexible in terms of variable selection by surgical teams. For example, depending on the magnitude of liver resection (i.e. minor or major), factors such as estimated blood loss (EBL), 3D visual clarity, tremor suppression, articulation, among others.

The robotic approach to performing a wide range of minimally invasive surgical procedures requires continuous feedback from surgical teams on conducting these procedures. That is, surgeons generally access active surgical robots remotely while manipulating robotic arms through joystick-like hand controllers to execute instructions or procedural steps.

As surgical robots become equipped with advanced sensors that can provide sensing capabilities while considering situations of surrounding entities (e.g. they become context-aware), they may predict procedural steps that are likely to be executed by surgical teams. Therefore, based on this context-awareness, they can calculate a risk factor with each possible procedural step. This can be helpful in cases surgeons are unable to assess a surgical step carefully or accurately. Because surgical robots are often equipped with advanced sensing and magnification capabilities, they can provide surgeons many perspectives from various angles during a surgical operation.

As part of our evaluation, we considered a surgical plan involving the Whipple procedure with an extended (right) hepatectomy. For the main Whipple procedure, we considered the decision variables: blood vessel removal or reconstruction rate, cancerous spread level (or risk), 3D imaging, among others. For the right hepatectomy, we considered a number of decision variables, such as estimated blood loss (EBL) rate, 3D visualization rate, distance from vein, among others. IoTSR allows surgeons to associate a priority with each decision variable representing the degree of importance of that variable on the overall decision making.

### A. Whipple Use Case

The Whipple procedure is a complex gastrointestinal operation to be performed using surgical robots because it involves three organs: (a) liver, pancreas, and intestines. We consider four main decision variables for this procedure as shown in Table I.

TABLE I
Whipple Procedure Use Case

| Priority (%) | Feature or Attribute |
|---|---|
| 10 | 3D visual clarity (%) |
| 10 | liver removal risk level (%) |
| 20 | blood vessel exposure or reconstruction (%) |
| 60 | cancerous spread level (%) |

Figures 3 and 4 present the results from executing the IoTSR trajectory planning optimization for the Whipple procedure use case. In this use case, we focus primarily on minimizing the likelihood operating on areas that are associated with high level of cancerous state and exposing blood vessels for removal or reconstruction.

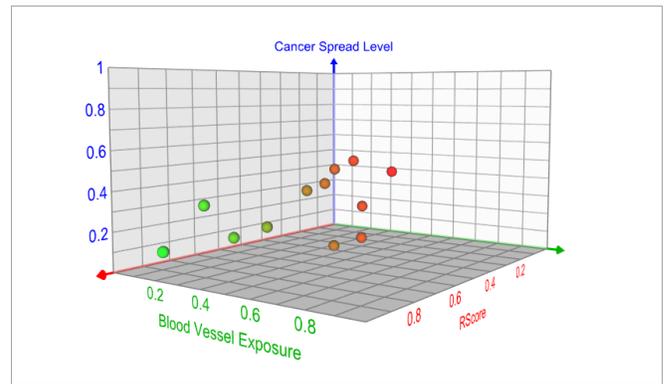

Fig. 3 IoTSR Trajectory Optimization for a Surgical Plan Involving Whipple Procedure Use Case (3D)

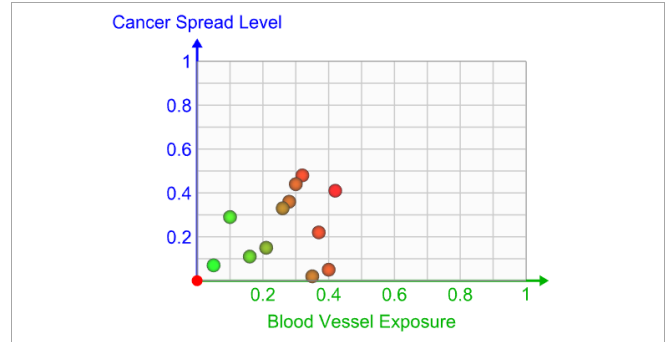

Fig. 4 Results from a Whipple Procedure Use Case (2D)

The green point in the graph shown in Figures 3 is associated with 0.07 cancer spread level (which we assume can be measured by the robot) and 0.05 blood vessel exposure (or construction) compared to the worst surgical step involving 0.41 and 0.409, respectively. In addition, IoTSR's optimization strategy using TOPSIS is capable of identifying a limited number of procedural steps, which indicates that can be used as an effective technique for surgeons when performing assistive robotic surgeries. By effectively assisting surgical teams using this self-adaptive IoT-based approach, the procedural time required can significantly be reduced.

### B. Hepatectomy Use Case

A surgical resection may require hepatectomy, the partial removal of the liver can be combined which can be with pancreaticoduodenectomy. This extension of the Whipple procedure can also be performed in a minimally invasion manner (e.g. robot-assisted) and vary in terms of the resection magnitude (e.g. major or minor). We focus primarily on the extension of the surgical plan involving right hepatectomy. We assume that surgeons have identified the patients that are approved or eligible for the type of hepatectomy procedure to be performed. IoTRS, which can reside on the robot itself or a computing device connected to the surgical robot, allows surgical teams to specify the decision variables to be employed and their level of importance on the overall outcome or decision-making process.

Table I1 presents the employed decision variables for this use case and the corresponding proprieties, which we assume are provided by a surgical team prior or in advance. Figures 5 and 6 present the results from executing the IoTSR optimization strategy based on the normalized values of all the decision variables considered for this use case (Table II).

TABLE II
Hepatectomy Use Case

| Priority (%) | Feature or Attribute |
|---|---|
| 10 | Jejunal mucosa (%) |
| 10 | bile duct diameter (mm) |
| 40 | estimated blood loss (EBL) (ml) |
| 40 | 3D visualization clarity (%) |

As presented in Figure 5, IoTSR clearly distinguishes between procedural steps that are optimal (shown in green) and those that are not recommended. For example, the green point (alternative 1) in Figures 5 and 6 is associated with an estimated blood loss rate of 0.03 and 0.52 3D visual clarity (VC) whereas the two red points (alternatives 11 and 12) are associated with 0.43 EBL, 0.06 VC and 0.30 EBL and 0.22 VC, respectively.

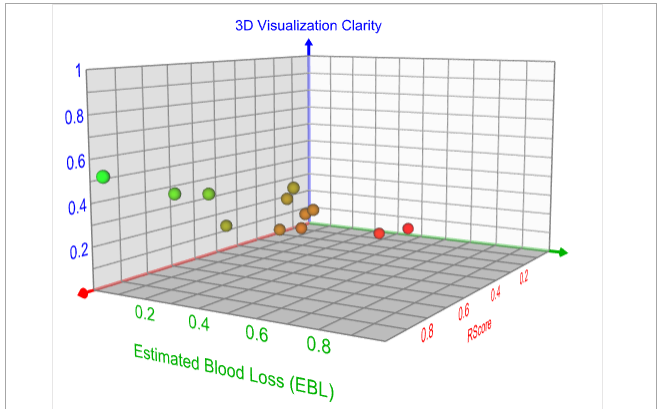

Fig. 5 IoTSR Trajectory Optimization for a Surgical Plan Involving Extension of the Hepatectomy Use Case

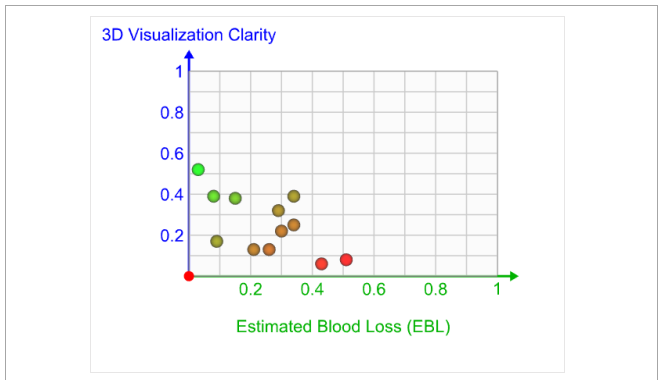

Fig. 6 Results from a Hepatectomy Extension Use Case (2D)

Because we would like to minimize the blood loss rate while maintaining a high-level of visualization clarity, it is ideal to select alternatives (or procedural steps) that accommodate the goals. Therefore, IoTSR optimally identifies alternative 1 as a recommended procedural step over procedural steps 11 and 12 which exhibit both very high blood loss rate and poor visualization clarity. IoTSR can then assist surgical teams by alerting them in case they select alternatives (or surgical procedure) that is not optimal. Because TOPSIS allows for weights to be adjusted, this provides a greater level of flexibility for surgeons to define their own preferences of the most important parameter which can vary throughout the entire surgical procedure.

## Conclusion

In this paper, we introduced an IoT-based surgical robot (IoTSR) framework that utilizes edge computing for assisting surgical teams during minimally invasive procedures involving robots. Throughout the paper, we presented a surgical plan that involves the Whipple procedure and extended through hepatectomy (or liver resection). We also identified possible decision variables that can be considered as part of our optimization strategy for recommending a surgical step as part of a surgical plan. We evaluated our IoTSR framework based on the proposed surgical plan and results show an effective self-adaptive optimization approach that can be employed as a tool for assisting surgical teams during robotic surgeries. For future work, we plan to incorporate additional surgical procedures and extend on the number of possible alternatives.